\documentclass[conference, 9pt]{IEEEtran}
\IEEEoverridecommandlockouts
\usepackage{cite}
\usepackage{amsmath,amssymb,amsfonts}
\usepackage{graphicx}
\usepackage{textcomp}
\usepackage{xcolor}
\usepackage[a4paper, total={184mm,239mm}]{geometry}
\def\BibTeX{{\rm B\kern-.05em{\sc i\kern-.025em b}\kern-.08em
    T\kern-.1667em\lower.7ex\hbox{E}\kern-.125emX}}
\usepackage{amsmath}
\usepackage{booktabs} 
\usepackage{multirow}
\usepackage{xcolor}
\usepackage[super]{nth}
\usepackage{tikz}
\newcommand*\circled[1]{\tikz[baseline=(char.base)]{
		\node[shape=circle,draw,inner sep=0.2pt] (char) {#1};}}
\newcommand*\circledB[1]{\tikz[baseline=(char.base)]{
            \node[shape=circle,fill,inner sep=0.2pt] (char) {\textcolor{white}{#1}};}}
\usepackage{soul}
\usepackage{enumitem}
\usepackage{algorithmic}
\usepackage{algorithm}

\usepackage{fancyhdr}
\fancypagestyle{firstpage}{
  \fancyhf{}
  \chead{To appear at the 58th IEEE/ACM Design Automation Conference (DAC), December 2021, San Francisco, CA, USA.}
  \cfoot{\thepage}
}

\usepackage[none]{hyphenat}

\begin{document}

\title{SpikeDyn: A Framework for Energy-Efficient Spiking Neural Networks with Continual and Unsupervised Learning Capabilities in Dynamic Environments
\vspace{-0.3cm}
}

\author{\IEEEauthorblockN{\textsuperscript{1}Rachmad Vidya Wicaksana Putra, \textsuperscript{2}Muhammad Shafique}
\IEEEauthorblockA{\textsuperscript{1}\textit{Technische Universit\"at Wien (TU Wien)}, Vienna, Austria \\
\textit{\textsuperscript{2}New York University Abu Dhabi (NYUAD)}, Abu Dhabi, United Arab Emirates \\
Email: rachmad.putra@tuwien.ac.at,
muhammad.shafique@nyu.edu}
\vspace{-0.8cm}
}
\maketitle
\thispagestyle{firstpage}

\begin{abstract}
Spiking Neural Networks (SNNs) bear the potential of efficient unsupervised and continual learning capabilities because of their biological plausibility, but their complexity still poses a serious research challenge to enable their energy-efficient design for resource-constrained scenarios (like embedded systems, IoT-Edge, etc.).
We propose SpikeDyn, a comprehensive framework for energy-efficient SNNs with continual and unsupervised learning capabilities in dynamic environments, for both the training and inference phases. 
It is achieved through the following multiple diverse mechanisms: 
1) reduction of neuronal operations, by replacing the inhibitory neurons with direct lateral inhibitions; 
2) a memory- and energy-constrained SNN model search algorithm that employs analytical models to estimate the memory footprint and energy consumption of different candidate SNN models and selects a Pareto-optimal SNN model; and 
3) a lightweight continual and unsupervised learning algorithm that employs adaptive learning rates, adaptive membrane threshold potential, weight decay, and reduction of spurious updates. 
Our experimental results show that, for a network with 400 excitatory neurons, our SpikeDyn reduces the energy consumption on average by 51\% for training and by 37\% for inference, as compared to the state-of-the-art.
Due to the improved learning algorithm, SpikeDyn provides on avg. 21\% accuracy improvement over the state-of-the-art, for classifying the most recently learned task, and by 8\% on average for the previously learned tasks. 
\end{abstract}

\begin{IEEEkeywords}
Spiking neural networks, SNNs, continual learning, unsupervised learning, energy-efficiency, embedded systems, complexity.
\vspace{-0.3cm}
\end{IEEEkeywords}

\vspace{-0.1cm}
\section{Introduction}
\label{Sec_Intro}
\vspace{-0.1cm}

With the evolution of neuromorphic computing, SNNs are (re-)gaining researchers' attention as they have demonstrated 
potential for having better learning capabilities (especially in unsupervised settings) due to their biological plausibility, and relatively better energy efficiency compared to other competitive neural network models \cite{Ref_Pfeiffer_DLSNN_FNINS18}. 
Previous works have explored different methodologies to build energy-efficient and unsupervised SNN systems, and most of them employed offline training \cite{Ref_Diehl_STDPmnist_FNCOM15, Ref_Srinivasan_EnhPlast_IJCNN17, Ref_Hazan_SOMSNN_IJCNN18, Ref_Saunders_LCSNN_NeuNet19, Ref_Putra_FSpiNN_TCAD20}.
However, the information learned by the offline-trained SNN system can be obsolete or may lead to low accuracy at run time under dynamically changing scenarios, as new data may have new features that should be learned online \cite{Ref_Panda_ASP_JETCAS18, Ref_Lobo_SNNonline_NeuNet19, Ref_Lesort_CLrobot_IF20, Ref_Anthes_LifelongLearn_ACM19, Ref_Ven_BIReplay_Nature20}. 
It becomes especially important for use cases like IoT-Edge devices deployed in dynamically changing environments \cite{Ref_Lobo_SNNonline_NeuNet19} and the robotic agents/UAVs in unknown terrains \cite{Ref_Lesort_CLrobot_IF20}. 
New data that are gathered directly from such dynamic environments, are usually unlabeled. 
Hence, an SNN-based system should employ unsupervised learning to process them \cite{Ref_Putra_FSpiNN_TCAD20}. 
Moreover, new data are uncontrolled and their classes might not be randomly distributed, thereby making the system difficult to learn different classes/tasks proportionally \cite{Ref_Allred_ForcedFiring_IJCNN16}. 
Therefore, the SNN system should employ \textit{real-time continual learning}\footnote{Continual learning is defined as the ability of a model to learn consecutive tasks (e.g., classes), while retaining information that have been learned 
\cite{Ref_Lobo_SNNonline_NeuNet19}\cite{Ref_Chen_Lifelong_MLP18}\cite{Ref_Parisi_CLL_NeuNet19}. Real-time means during the operational lifetime of the dynamic system.}, while avoiding the undesired conditions, such as:
(1) the system learns information from the new data, but quickly forgets the previously learned ones (i.e., \textit{catastrophic forgetting}) 
\cite{Ref_McCloskey_CI_Elsevier89, Ref_Chen_Lifelong_MLP18, Ref_Parisi_CLL_NeuNet19}; 
(2) the system mixes new information with the existing ones, thereby corrupting/polluting the existing information \cite{Ref_Panda_ASP_JETCAS18}\cite{Ref_Allred_ForcedFiring_IJCNN16}; and  
(3) the learning requires a large number of weights and neuron parameters, and complex exponential calculations, thereby consuming high energy.

\smallskip
\textbf{Targeted Research Problem:} 
\textit{If and how can we design lightweight and energy-efficient continual and unsupervised learning for SNNs that adapts to task changes (dynamic environments) for providing improved accuracy at run time}. 
An efficient solution to this problem will enable SNN systems to achieve better learning capabilities on energy-constrained devices deployed in the unpredictable dynamic scenarios.

\smallskip
\textbf{State-of-the-Art and Limitations:}
Previous works have tried to achieve continual learning through different techniques. 
First category includes \textit{the supervised learning techniques} that minimize the cost function in the learning process using data labels \cite{Ref_Kirkpatrick_EWC_PNAS17,Ref_Lee_OvercomeCF_NIPS17,Ref_Wysoski_OLstructural_ICANN06}. 
Hence, they cannot process unlabeled data which is required in the targeted problem.
Second category includes \textit{the unsupervised learning techniques} that perform learning using unlabeled data  \cite{Ref_Panda_ASP_JETCAS18}\cite{Ref_Allred_ForcedFiring_IJCNN16}\cite{Ref_Allred_CFN_FNINS20}. 
However, they suffer from spurious updates which lead to the sub-optimal accuracy, since they update the weights at each spike event, as observed in \cite{Ref_Srinivasan_EnhPlast_IJCNN17}.
They also incur high energy consumption due to: 
(a) additional neurons \cite{Ref_Allred_ForcedFiring_IJCNN16}\cite{Ref_Allred_CFN_FNINS20};
(b) non-proportional quantities of training samples, i.e., samples from the earlier task are presented with larger quantities than later tasks \cite{Ref_Panda_ASP_JETCAS18};
and (c) large memory footprint and complex exponential calculations to achieve high accuracy \cite{Ref_Panda_ASP_JETCAS18}, i.e., a total of 800 neurons are needed to achieve 75\% accuracy on the MNIST dataset. Note, the research for the continual and unsupervised learning in SNNs is still at an early stage and prominent SNN works mostly use the MNIST dataset \cite{Ref_Panda_ASP_JETCAS18}
\cite{Ref_Allred_ForcedFiring_IJCNN16}
\cite{Ref_Allred_CFN_FNINS20}. Therefore, we adopt the same test conditions as used widely by the SNN research community.
\textit{To highlight the targeted problem and the limitations of the state-of-the-art, we perform an experimental case study using the MNIST dataset, as discussed below}.

\vspace{-0.1cm}
\subsection{Motivational Case Study}
\label{Sec_CaseStudy}
\vspace{-0.1cm}

\begin{figure}[t]
\centering
\includegraphics[width=\linewidth]{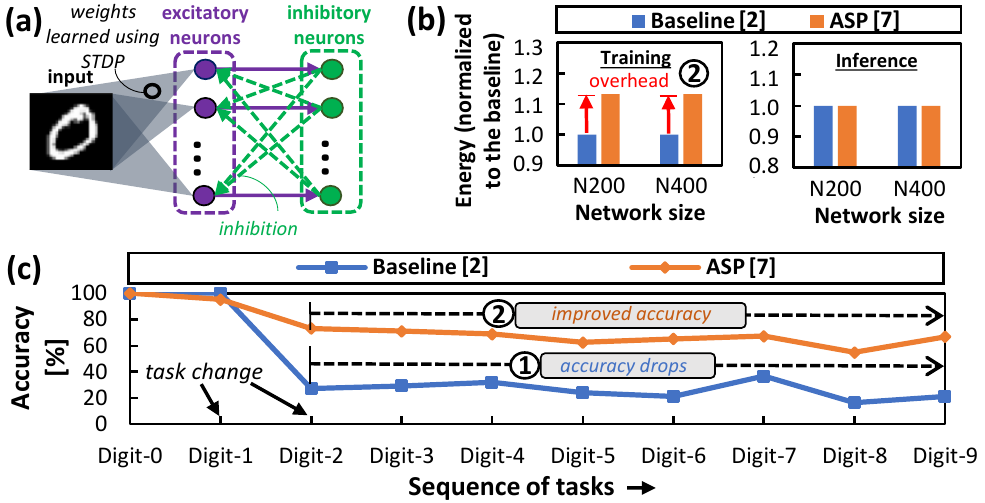}
\vspace{-0.7cm}
\caption{(a) The SNN architecture employed by the baseline \cite{Ref_Diehl_STDPmnist_FNCOM15} and the state-of-the-art (ASP) \cite{Ref_Panda_ASP_JETCAS18}. 
The experimental results: (b) the energy consumption for networks with 200 excitatory neurons (N200) and 400 excitatory neurons (N400), and (c) the per-digit accuracy for N400.}
\label{Fig_ObserveSoA}
\vspace{-0.5cm}
\end{figure}

We performed experiments that provide dynamic scenarios by feeding consecutive task (i.e., class\footnote{The SNN community uses term ``task" which refers to ``class" in dataset.}) changes to the network. 
First, a stream of samples for digit-0 is fed. 
Then, the task is changed to digit-1. 
This process is repeated for other tasks without re-feeding previous tasks, and each task has the same number of samples.
More details of the experimental setup are presented in Section~\ref{Sec_EvalMethod}.
The results are presented in Fig.~\ref{Fig_ObserveSoA}, from where we make the following key observations.
\circled{1}~The baseline does not efficiently learn new tasks from digit-2 onward, as most of the synapses are already occupied by previously learned tasks (digit-0 and digit-1), and mix new information with the existing ones.
\circled{2}~The state-of-the-art improves the accuracy over the baseline at a cost of an energy overhead due to: 
(a) a large number of weights and neuron parameters from excitatory and inhibitory layers, and 
(b) the complex exponential calculations for computing multiple spike traces, membrane-potential and threshold-potential decay, and weight decay.
\textit{These observations expose several challenges that need to be solved to address the targeted problem, as discussed below.}

\subsection{Key Scientific Challenges that are Addressed in this Paper}
\label{Sec_Challenges}

\begin{itemize}[leftmargin=*]
    \item The SNN system should employ a simple yet effective learning algorithm at run time that achieves high accuracy, in both the dynamic and non-dynamic scenarios.
    \item It should reduce the non-significant weights or neuron parameters, and the complex exponential calculations, to optimize the energy consumption. 
    \item The memory- and energy-constraints should be considered in the design process to meet many design scenarios.
\end{itemize}

\subsection{Our Novel Contributions and Key Results}
\label{Sec_NovelContribute}

\begin{figure}[hbtp]
\centering
\includegraphics[width=\linewidth]{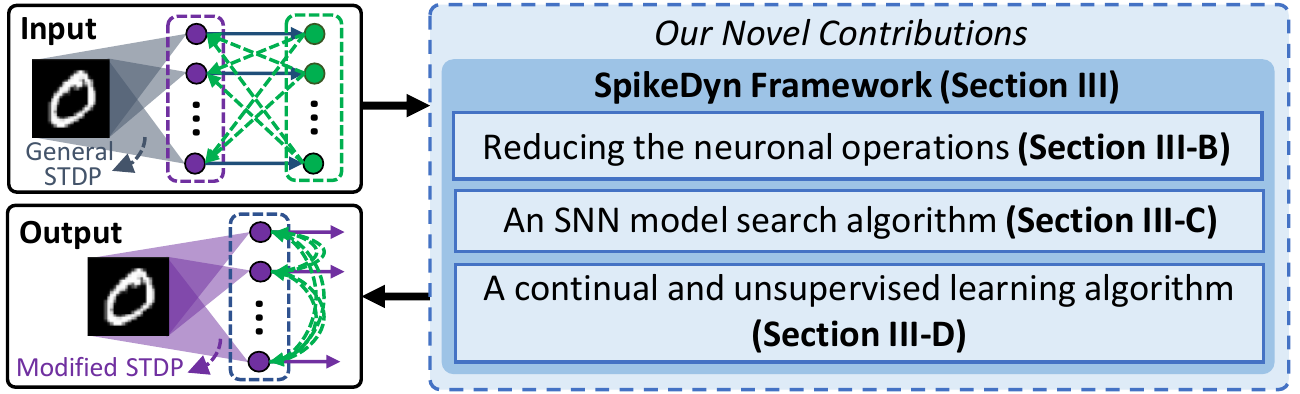}
\vspace{-0.6cm}
\caption{The overview of novel contributions (shown in the blue boxes).}
\label{Fig_NovelContributions}
\vspace{-0.5cm}
\end{figure}

To address the above challenges, we propose \textit{SpikeDyn framework (Section~\ref{Sec_SpikeDynFrame})} 
for developing energy-efficient SNNs considering both the training and the inference phases, with continual and unsupervised learning capabilities in dynamic environments. 
The SpikeDyn employs the following key mechanisms (see Fig.~\ref{Fig_NovelContributions}).
\begin{enumerate}[leftmargin=*]
    \item \textit{Reducing the energy consumption of the neuronal operations (Section~\ref{Sec_SpikeDyn_ReduceNeuron})} by replacing the inhibitory neurons with the direct lateral inhibitory connections.
    \item \textit{An SNN model search algorithm (Section~\ref{Sec_SpikeDyn_DSEalgorithm})} under the given memory- and energy-constraints. 
    It quickly estimates the memory footprint and energy consumption of the investigated SNN models using our analytical models that leverage the network parameters, the bit precision, the energy for processing an input, and the number of samples.
    \item \textit{An algorithm for achieving a continual and unsupervised learning (Section~\ref{Sec_SpikeDyn_ContinualAlg})} through the following means: (a) adaptive learning rates, (b) synaptic weight decay, (c) adaptive membrane threshold potential, and (d) reduction of the spurious weight updates.
\end{enumerate}

\smallskip
\textit{Key Results:}
We evaluated our SpikeDyn for accuracy and energy consumption using Python-based simulations with the MNIST on the Embedded GPU and the GPGPUs, to show the generality of our solution. 
For a network with 400 excitatory neurons, when compared to the state-of-the-art \cite{Ref_Panda_ASP_JETCAS18}, SpikeDyn reduces the energy consumption by up to 66\% (avg. 51\%) for training and up to 54\% (avg. 37\%) for inference. 
It also improves the accuracy by up to 29\% (avg. 21\%) for the most recently learned task and by up to 37\% (avg. 8\%) for the previously learned tasks, in dynamic scenarios.

\renewcommand{\headrulewidth}{0pt}
\section{Background: An Overview of SNNs}
\label{Sec_Background}

SNNs are considered as the third generation of neural networks' computation models, as they exhibit high biological plausibility through their spike-coded information. 
An SNN model consists of \textit{spike coding}, \textit{network architecture}, \textit{neuron and synapse models}, and \textit{learning rule} \cite{Ref_Tavanaei_DLSNN_Neunet18}. 
Spike coding converts the information into a spike sequence/train. 
There are several coding schemes in literature, such as: rate, temporal, rank-order, phase, and burst coding schemes \cite{Ref_Gautrais_SpikeCoding_Bio98, Ref_Kayser_PhaseCoding_Neuron09, Ref_Thorpe_RankOrder_Springer98,Ref_Park_BurstSNN_DAC19}.
Here, we consider rate coding as it has demonstrated high accuracy in unsupervised SNNs \cite{Ref_Diehl_STDPmnist_FNCOM15}. 
We also consider the network architecture of Fig.~\ref{Fig_ObserveSoA}(a), since it is widely employed in the literature for the continual and unsupervised learning-based SNNs \cite{Ref_Panda_ASP_JETCAS18}\cite{Ref_Allred_ForcedFiring_IJCNN16}. 
Here, each excitatory neuron is expected to recognize a class. 
For the neuron model, we use the Leaky Integrate-and-Fire model, since it has the lowest computational complexity among the existing neuron models. 
This model increases its membrane potential at the occurrence of the incoming spike. 
It generates a spike when the membrane potential reaches the threshold potential ($V_{th}$), and then goes to the reset potential ($V_{reset}$). 
To prevent a neuron from dominating the spiking activity, the membrane threshold potential is usually defined by $V_{th}+\theta$ \cite{Ref_Diehl_STDPmnist_FNCOM15}.  
The $\theta$ denotes the adaptation potential which is increased each time the neuron spikes, otherwise it decays with a rate of $\theta_{decay}$.
Meanwhile, a synapse is modeled by the conductance and weight ($w$). 
The conductance is increased by $w$ when a spike arrives at a synapse, otherwise it decays exponentially. 
For the learning rule, the SNN model employs a spike-timing-dependent plasticity (STDP) mechanism. 
Further details on SNNs can be found in \cite{Ref_Pfeiffer_DLSNN_FNINS18}\cite{Ref_Tavanaei_DLSNN_Neunet18}.

\section{SpikeDyn Framework}
\label{Sec_SpikeDynFrame}
\vspace{-0.2cm}

\subsection{Overview}
\vspace{-0.1cm}

\begin{figure}[hbtp]
\vspace{-0.3cm}
\centering
\includegraphics[width=\linewidth]{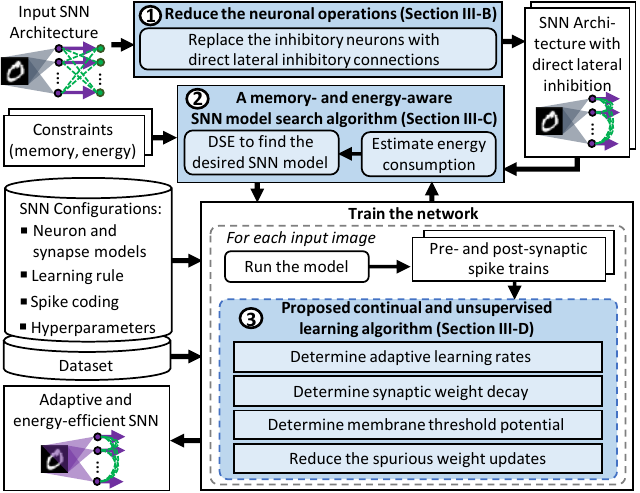}
\vspace{-0.6cm}
\caption{Overview of our SpikeDyn framework; see novel blocks in blue.}
\label{Fig_SpikeDyn}
\vspace{-0.4cm}
\end{figure}

Fig.~\ref{Fig_SpikeDyn} shows the detailed flow of our SpikeDyn framework with the following key steps, which are explained in the subsequent sections.

\begin{enumerate}[leftmargin=*]
    \item \textbf{Reducing the operations in SNN model} to minimize the energy consumption, through a replacement of the inhibitory neurons with the direct lateral inhibitions. Thus, the operations in the inhibitory layer are eliminated. 
    \item \textbf{An SNN model search algorithm} that explores different number of excitatory neurons to find a set of Pareto-optimal SNN models, while considering the memory and energy constraints.
    To quickly perform the search, we also propose analytical models that achieve less than 5\% errors compared to the actual run, by incorporating: 
    \begin{enumerate}[leftmargin=*]
        \item the number of weights and neuron parameters, and the bit precision, to estimate the memory footprint, 
        \item the energy consumption for processing an input sample, and the number of samples that will be processed, to estimate the energy consumption.
    \end{enumerate}
    \item \textbf{A continual and unsupervised learning algorithm} that employs the following techniques. 
    \begin{enumerate}[leftmargin=*]
        \item \textit{Adaptive learning rates} that determine the potentiation and depression factors in the STDP-based learning, using the presynaptic and postsynaptic spike activities.   
        \item \textit{Synaptic weight decay} that helps the network to gradually remove the weak synaptic connections (which represent old and insignificant information), thereby enabling the synapses to learn new information. 
        \item \textit{Adaptive membrane threshold potential} that provides balance in the neurons' internal, so that the neuron generates spikes only when the corresponding synapses need to learn the input features.  
        It is determined by the threshold potential value and its decay rate.  
        \item \textit{Reduction of the spurious weight updates} in the presynaptic and postsynaptic spike event by employing timestep to carefully perform weight potentiation and depression. 
    \end{enumerate}
\end{enumerate}

\subsection{Reducing the neuronal operations}
\label{Sec_SpikeDyn_ReduceNeuron}
\vspace{-0.1cm}

\begin{figure}[hbtp]
\vspace{-0.3cm}
\centering
\includegraphics[width=\linewidth]{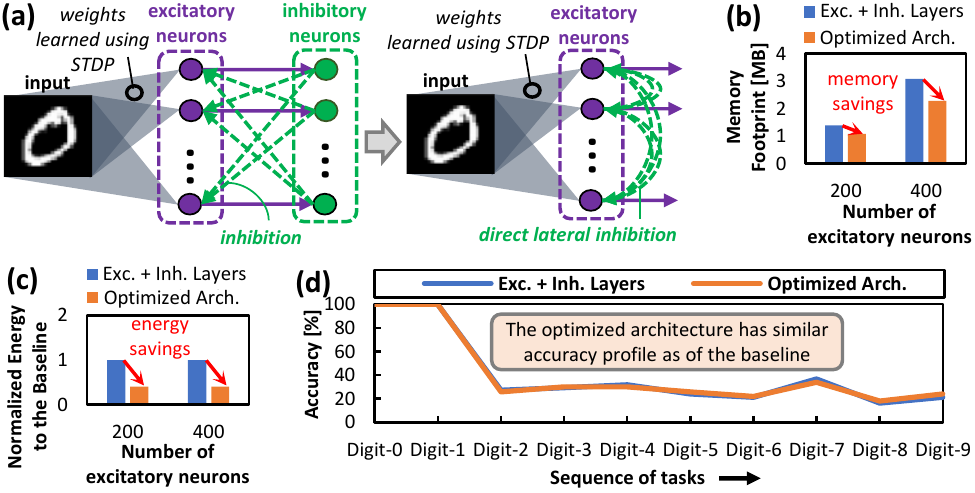}
\vspace{-0.6cm}
\caption{(a) Replacing the inhibitory neurons with the direct lateral inhibitions. 
The optimized architecture reduces (b) memory and (c) energy, but still has similar (d) accuracy profile in dynamic scenarios like the baseline architecture.}
\label{Fig_SpikeDynReduceNeuron}
\vspace{-0.3cm}
\end{figure}

Previous works in the continual and unsupervised learning for SNNs, used the network architecture shown in Fig.~\ref{Fig_ObserveSoA}(a), which consists of the input, excitatory, and inhibitory layers \cite{Ref_Panda_ASP_JETCAS18}\cite{Ref_Allred_ForcedFiring_IJCNN16}. 
We observe that the inhibitory neurons have different parameters from the excitatory ones to be saved in memory. 
Therefore, employing such an architecture will consume high memory and energy.
To address this issue, we reduce the neuronal operations to substantially decrease the memory footprint and the energy consumption, as shown in Figs.~\ref{Fig_SpikeDynReduceNeuron}(a)-\ref{Fig_SpikeDynReduceNeuron}(c).
We also observe that, the optimized architecture still achieves similar accuracy profile as of the baseline, as shown in Fig.~\ref{Fig_SpikeDynReduceNeuron}(d). 
Therefore, we will improve the accuracy of the optimized architecture with our learning mechanism, as discussed in the Section~\ref{Sec_SpikeDyn_ContinualAlg}.

\vspace{-0.1cm}
\subsection{An SNN model search algorithm}
\label{Sec_SpikeDyn_DSEalgorithm}
\vspace{-0.1cm}

Each application has memory- and energy-constraints that need to be considered in the model generation.
Towards this, we propose an algorithm to search for an appropriate model size for a given SNN architecture that meets the design constraints (see Alg.~\ref{Alg_DSE}). 
The idea is to explore different sizes of SNN model and estimate their memory and energy consumption in both training and inference phases, using the proposed analytical models. 
%
\begin{algorithm}[hbtp]
\scriptsize
\caption{Pseudo-code of the proposed search algorithm}
\label{Alg_DSE}
\begin{algorithmic}[1]
\renewcommand{\algorithmicrequire}{\textbf{INPUT:}}
\renewcommand{\algorithmicensure}{\textbf{OUTPUT:}}
\REQUIRE \textbf{(1)} Memory constraint ($mem_c$); 
\textbf{(2)} Energy constraints: training ($E_{ct}$), inference ($E_{ci}$); 
\textbf{(3)} SNN model ($model$): number of neurons ($model.n_{exc}$), size ($model.mem$), energy of training ($model.E_t$) and inference ($model.E_i$);
\textbf{(4)} Energy for one sample: training ($E_{1t}$), inference ($E_{1i}$); 
\textbf{(5)} number of additional neurons ($n_{add}$); \\
\ENSURE SNN model ($model$); \\
\vspace{0.1cm}
\renewcommand{\algorithmicrequire}{\textbf{BEGIN}}
\renewcommand{\algorithmicensure}{\textbf{END}}
\REQUIRE \hspace{0.1cm} \\   
    \textbf{Initialization}: \\
      \STATE $model.n_{exc} = 0$; \\
      \STATE $model.mem = 0$; \\
    \textbf{Process}: \\
      \WHILE{$model.mem \leq mem_c$}
        \IF{($model.n_{exc} > 0$)}
          \STATE perform \textit{training} with 1 sample using Alg. \ref{Alg_ContinualAlg}; \\
          \STATE calculate $E_{1t}$; // for 1 sample \\
          \STATE estimate $model.E_t$; // for all samples \\
          \IF{($model.E_t \leq E_{ct})$}
            \STATE perform \textit{inference} with 1 sample; \\
            \STATE calculate $E_{1i}$; // for 1 sample \\
            \STATE estimate $model.E_i$; // for all samples \\
            \IF{($model.E_i \leq E_{ci}$)} 
              \STATE save $model$; \\
            \ENDIF 
          \ENDIF 
        \ENDIF 
        \STATE $model.n_{exc} += n_{add}$;
        \STATE estimate $model.mem$;
      \ENDWHILE 
    \RETURN $model$; \\
\ENSURE 
\end{algorithmic} 
\end{algorithm}
\setlength{\textfloatsep}{4pt}
\vspace{-0.2cm}

\begin{figure}[hbtp]
\centering
\includegraphics[width=\linewidth]{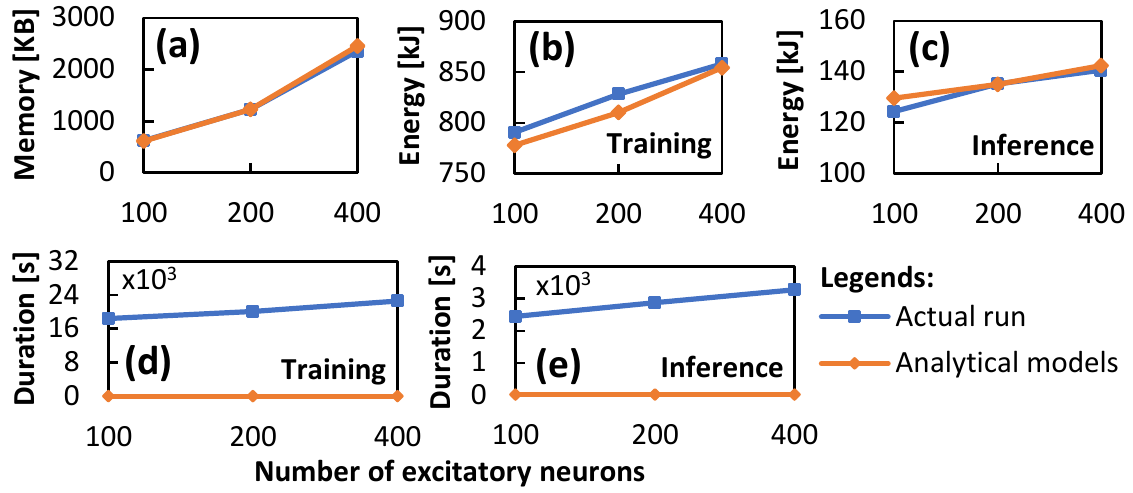}
\vspace{-0.7cm}
\caption{Validation of our analytical models against the actual runs in terms of (a) the memory footprint, and the energy consumption for (b) training and (c) inference, using full MNIST dataset. Our algorithm also reduces the exploration time over the actual runs for both (d) training and (e) inference phases.}
\label{Fig_Benefit_DSE}
\vspace{-0.1cm}
\end{figure}

For each investigated SNN models' size, the memory footprint ($mem$) is estimated using $ mem = (P_{w} + P_{n}) \cdot BP$, that leverages the number of weights ($P_w$) and neuron parameters ($P_n$), and the bit precision ($BP$).
The reason is that, the above aspects are dominant factors in determining the size of an SNN model. 
Meanwhile, the total energy ($E$) is estimated using $E = E_1 \cdot N$, that leverages the energy for processing a single sample ($E_1$), and the number of samples that will be processed ($N$).
The number of samples $N$ is important, as the deployed systems might have different number of samples available from the environment.
If the estimated memory $mem$ and energy $E$ are within the memory constraint ($mem_c$) and energy constraint ($E_c$) respectively, then the investigated SNN model is selected as the candidate of solution.
Our algorithm then selects the largest-sized SNN model from the candidates as the solution, since larger network usually can achieve higher accuracy \cite{Ref_Putra_FSpiNN_TCAD20}.
We validated our analytical models against the actual execution runs, see the results presented in Fig.~\ref{Fig_Benefit_DSE}(a) for memory footprint, and Figs.~\ref{Fig_Benefit_DSE}(b)-\ref{Fig_Benefit_DSE}(c) for energy consumption of training and inference, respectively. 
The results show that our analytical models achieve less than 5\% errors compared to the actual runs. 
Thus, they are suitable to the fast estimation need.
Employing our algorithm is beneficial, rather than actually running all possible SNN configurations and selecting the desired one at the end, since it saves the exploration time, as shown in Figs.~\ref{Fig_Benefit_DSE}(d)-\ref{Fig_Benefit_DSE}(e).

\subsection{Proposed continual and unsupervised learning algorithm}
\label{Sec_SpikeDyn_ContinualAlg}

Our SpikeDyn employs an algorithm that incorporates the following techniques (see pseudo-code in Alg.~\ref{Alg_ContinualAlg}).

\smallskip
\textbf{Adaptive Learning Rates:}
Our algorithm employs the potentiation factor ($k_p$) and the depression factor ($k_d$) to define the learning rates for weight potentiation and depression. 
The idea is adjusting the potentiation factor $k_p$ to have high value when the corresponding synapses need to learn input features, which is indicated by the occurrences of postsynaptic spikes. 
The value of $k_p$ is obtained by normalizing the maximum accumulated postsynaptic spikes ($maxSp_{post}$) with the spike threshold ($Sp_{th}$); see Eq.~\ref{Eq_STDP_k}(a).
Meanwhile, the depression factor $k_d$ provides weight depression when the corresponding synapses need to weaken the connections, which is indicated by no occurrences of postsynaptic spikes.
The value of $k_d$ is obtained by the ratio between the maximum accumulated postsynaptic spikes ($maxSp_{post}$) and presynaptic spikes ($maxSp_{pre}$); see Eq.~\ref{Eq_STDP_k}(b).
These factors are incorporated into the improved STDP-based learning algorithm; see Eq.~\ref{Eq_STDP_Improved}.
Here, $\Delta w$ denotes the weight change, $\eta_{pre}$ and $\eta_{post}$ denote the learning rate for a pre- and post-synaptic spike, $x_{pre}$ and $x_{post}$ denote the pre- and post-synaptic traces, respectively.

\vspace{-0.1cm}
\begin{equation}
\vspace{-0.3cm}
\small
\textbf{(a)} \; \;  k_p = \left \lceil \frac{maxSp_{post}}{Sp_{th}} \right \rceil \; \; ; \; \; \textbf{(b)} \; \; k_d = \frac{maxSp_{post}}{maxSp_{pre}} \\
\label{Eq_STDP_k}
\vspace{-0.2cm}
\end{equation}

\begin{equation}
\small
\begin{split}
\Delta w = 
\begin{cases}
-k_d \cdot \eta_{pre} \cdot  x_{post} & \text{on} \; \text{depression update time}\\
k_p \cdot \eta_{post} \cdot x_{pre} & \text{on} \; \text{potentiation update time}
\end{cases}
\label{Eq_STDP_Improved}
\end{split}
\vspace{-0.1cm}
\end{equation}

\begin{figure}[t]
\centering
\includegraphics[width=\linewidth]{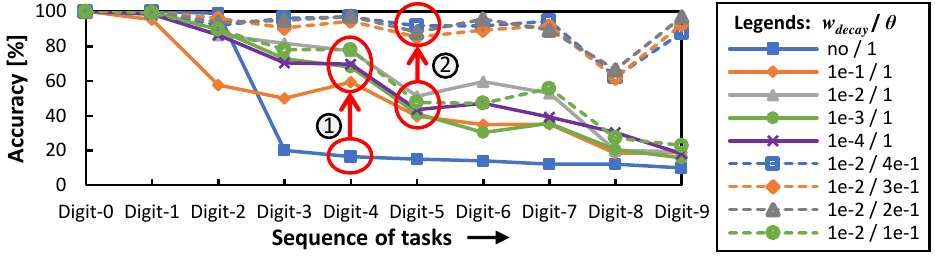}
\vspace{-0.6cm}
\caption{Impact of employing weight decay and adaptation potential $\theta$ on the accuracy of learning new tasks in a dynamic scenario.}
\label{Fig_ObserveThetaWdecay}
\vspace{-0.1cm}
\end{figure}

\textbf{Synaptic Weight Decay:}
We employ a weight decay to gradually remove the old and insignificant information, which are represented by small weight values.
It follows equation $\tau_{decay} \cdot (dw/dt) = -w_{decay} \cdot w$, with $\tau_{decay}$ denotes the decay time constant and $w_{decay}$ denotes the weight decay rate. 
In this manner, weak connections will get more disconnected over the training period.
We define the value of $w_{decay}$ to be inversely proportional to the size of network ($w_{decay} \propto 1/n_{exc}$), with $n_{exc}$ denotes the number of excitatory neurons. 
The reason is that, a smaller network has less number of synapses for learning new information, while retaining the old ones.
Therefore, it needs to forget the old information at a faster rate than the larger network. 
We observe that an appropriate $w_{decay}$ can improve accuracy, as shown by the label-\circled{1} in Fig.~\ref{Fig_ObserveThetaWdecay}.

\smallskip
\textbf{Adaptive Membrane Threshold Potential:}
The threshold potential is defined by $V_{th}+\theta$, as discussed in Section~\ref{Sec_Background}.
We observe that the adaptation potential $\theta$ has an important role to determine whether a neuron would generate spikes easily for later inputs.
If $\theta$ is too high, the neurons will not spike easily for later inputs, since the threshold potential is already adjusted for recognizing earlier inputs. 
In the context of dynamic scenarios, the network will face difficulties when learning new tasks.
If $\theta$ is too low, the neurons will spike easily for any inputs. 
In the context of dynamic scenarios, the network will quickly forget old information.
Thus, the threshold potential should be balanced, so that some neurons are available for recognizing new features, while the others retain the old yet significant information. 
Towards this, we define the adaptation potential $\theta$ to be proportional to its decay rate $\theta_{decay}$ and the presentation time of a sample $t_{sim}$, and can be stated as $\theta = c_{\theta} \cdot \theta_{decay} \cdot t_{sim}$, with $c_{\theta}$ denotes the adaptation constant.
An appropriate $\theta$ can improve accuracy, as shown by the label-\circled{2} in Fig.~\ref{Fig_ObserveThetaWdecay}.

\begin{figure}[hbtp]
\vspace{-0.2cm}
\centering
\includegraphics[width=0.95\linewidth]{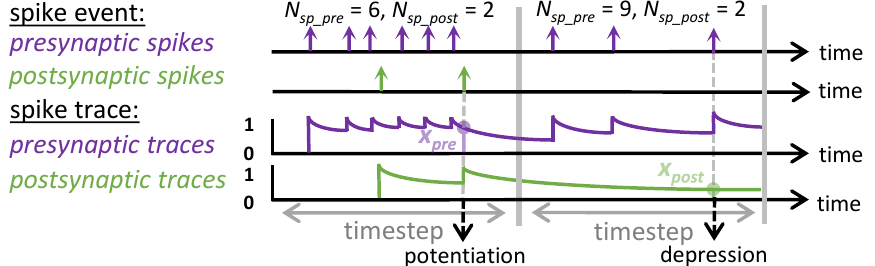}
\vspace{-0.2cm}
\caption{Overview of the proposed timestep-based weight updates.}
\label{Fig_SpikeDyn_Timestep}
\vspace{-0.2cm}
\end{figure}

\smallskip
\textbf{Reducing the Spurious Weight Updates:}
Previous work \cite{Ref_Srinivasan_EnhPlast_IJCNN17} has observed that there are spurious updates in SNNs, which can degrade the accuracy. 
These are observed in two cases: 
(1) when the neurons spike unpredictably, due to the random weight initialization; and 
(2) when a neuron spikes for patterns that belong to different classes, due to the overlapped features.
We exploit this observation in a novel way to reduce the spurious updates that are induced by both the pre- and post-synaptic spikes.
The idea is to employ a timestep, and then monitor whether at least one postsynaptic spike happens. If so, then the weight potentiation will be conducted, and otherwise, the weight depression will be conducted (see Fig.~\ref{Fig_SpikeDyn_Timestep}). 

\begin{algorithm}[t]
\scriptsize
\caption{\color{black} Pseudo-code of the proposed learning algorithm}
\label{Alg_ContinualAlg}
\begin{algorithmic}[1]
\renewcommand{\algorithmicrequire}{\textbf{INPUT:}}
\renewcommand{\algorithmicensure}{\textbf{OUTPUT:}}
\REQUIRE 
\textbf{(1)} Simulation time for an input ($t_{sim}$); 
\textbf{(2)} Timestep ($t_{step}$); 
\textbf{(3)} SNN parameters: \# of neurons ($n_{exc}$), \# of synapses-per-neuron ($n_{syn}$), accumulated presynaptic spikes ($N_{sp\_pre}$) and accumulated postsynaptic spikes ($N_{sp\_post}$); 
\textbf{(4)} Presynaptic spike ($sp_{pre}$), postsynaptic spike ($sp_{post}$);\\
\ENSURE Synaptic weight update ($\Delta w$); \\
\vspace{0.1cm}
\renewcommand{\algorithmicrequire}{\textbf{BEGIN}}
\renewcommand{\algorithmicensure}{\textbf{END}}
\REQUIRE \hspace{0.1cm} \\   
    \textbf{Initialization}: \\
      \STATE $\Delta w[n_{exc}, n_{syn}] = zeros[n_{exc},n_{syn}]$; \\
      \STATE $N_{sp\_pre}[n_{exc}, n_{syn}] = zeros[n_{exc}, n_{syn}]$; \\
      \STATE $N_{sp\_post}[n_{exc}] = zeros[n_{exc}]$; \\
    \textbf{Process}: \\
        \FOR{$(t = 0$ to $(t_{sim}-1))$}
          \FOR{$(i = 0$ to ($n_{exc}-1))$}
            \FOR{$(j = 0$ to ($n_{syn}-1))$}
              \IF{$sp_{pre}$}
                \STATE $N_{sp\_pre}[i,j]$ +$=$ $1$;
              \ENDIF
              \IF{$sp_{post}$}
                \STATE $N_{sp\_post}[i]$ +$=$ $1$;
              \ENDIF
            \ENDFOR
          \ENDFOR
          \IF{$((t \mod t_{step})$ == $0)$}
            \STATE $maxSp_{pre} = max(N_{sp\_pre})$; 
            \STATE $maxSp_{post} = max(N_{sp\_post})$;
            \IF{(no $sp_{post}$ within $t_{step})$}
              \STATE update $\Delta w[:,:]$ using Eq.~\ref{Eq_STDP_Improved}; // weight depression
            \ELSE
              \STATE $m \leftarrow index(max(N_{sp\_post}))$;
              \STATE update $\Delta w[m,:]$ using Eq.~\ref{Eq_STDP_Improved}; // weight potentiation 
            \ENDIF
          \ELSE
            \STATE update $\Delta w[:,:]$ using weight decay;   
          \ENDIF
          \RETURN $\Delta w$; 
        \ENDFOR
\ENSURE 
\end{algorithmic}
\end{algorithm}
\setlength{\textfloatsep}{2pt}

\vspace{-0.2cm}
\section{Evaluation Methodology}
\label{Sec_EvalMethod}
\vspace{-0.1cm}

\begin{figure}[hbtp]
\vspace{-0.2cm}
\centering
\includegraphics[width=\linewidth]{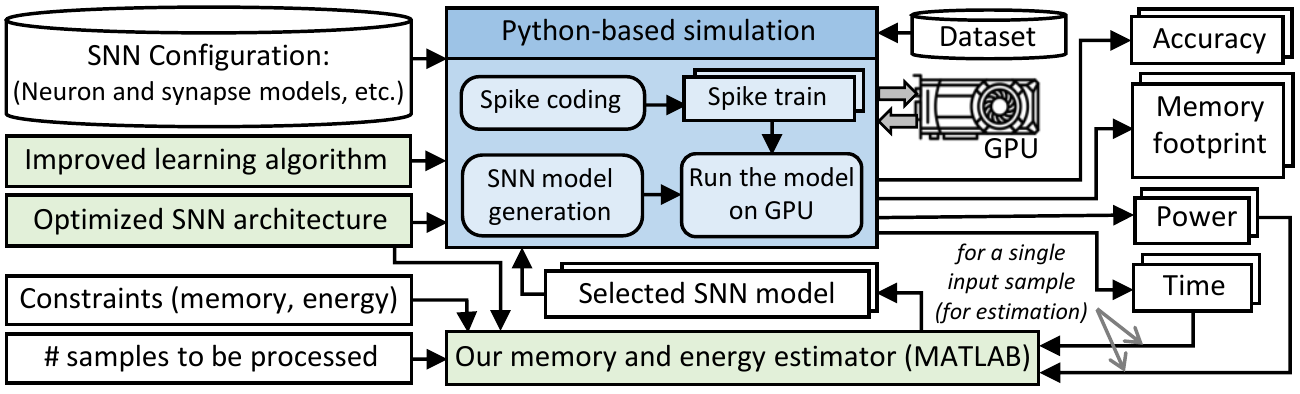}
\vspace{-0.6cm}
\caption{The experimental setup and tool flow.}
\label{Fig_ExpSetup}
\vspace{-0.2cm}
\end{figure}

Fig.~\ref{Fig_ExpSetup} shows the experimental setup for evaluating SpikeDyn framework.   
We used Python-based SNN simulations \cite{Ref_Hazan_BindsNET_FNINF18} that run on Embedded GPU (Nvidia Jetson Nano) and GPGPUs (Nvidia GTX 1080 Ti and RTX 2080 Ti) to perform diverse evaluations under different memory and compute capabilities, for showing the generality of our solution. 
The GPU specifications are presented in Table~\ref{Table_GPUspecs}.  

\vspace{-0.2cm}
\begin{table}[hbtp]
\vspace{-0.2cm}
\caption{GPU Specifications.}
\label{Table_GPUspecs}
\vspace{-0.2cm}
\centering
\scriptsize
\begin{tabular}{|c|c|c|c|}
\hline 
\textbf{Category} & \textbf{Jetson Nano} & \textbf{GTX 1080 Ti} & \textbf{RTX 2080 Ti} \\
\hline
\hline
Architecture & Maxwell & Pascal & Turing \\
\hline
CUDA cores & 128 & 3584 & 4352 \\
\hline
Memory & 4GB LPDDR4 & 11GB GDDR5X & 11GB GDDR6 \\
\hline
Interface width & 64-bit & 352-bit & 352-bit \\
\hline
Power & 10W & 250W & 250W\\
\hline
\end{tabular}
\vspace{-0.2cm}
\end{table}

To estimate the energy consumption of both the training and the inference phases, we adopted the approach of \cite{Ref_Han_DeepCompress_ICLR16}. 
That is, leveraging the information of the processing time, and the processing power that is reported using (1) \textit{nvidia-smi} utility for GPGPUs, and (2) measurement using a power-meter for Embedded GPU.
We used the MNIST as it is widely used for evaluating the continual and unsupervised learning in SNNs \cite{Ref_Panda_ASP_JETCAS18}\cite{Ref_Allred_ForcedFiring_IJCNN16}\cite{Ref_Allred_CFN_FNINS20}, and employed the rate coding to convert each pixel of an image into a Poisson-distributed spike train. 
For comparison partners, we used work in \cite{Ref_Diehl_STDPmnist_FNCOM15} as the baseline, and the adaptive synaptic plasticity (ASP) \cite{Ref_Panda_ASP_JETCAS18} as the state-of-the-art\footnote{We used the work in \cite{Ref_Panda_ASP_JETCAS18} since it is the only available recent work that has the complete set of configurations, parameters, and implementations details to have a reproducible design and results.}.
The evaluation was performed for both dynamic and non-dynamic environments. 
Dynamic environments mean that the network is fed with consecutive task changes without re-feeding previous tasks, and each task has the same number of samples.
It simulates an extreme condition where an SNN system receives training tasks from the environment in a consecutive manner, and each task has a defined number of samples.
Non-dynamic environments mean that the network is fed with input samples whose tasks are distributed randomly. 
We used different network sizes, i.e., 200 and 400 excitatory neurons, which we refer them to as N200 and N400, respectively. 

\section{Experimental Results and Discussions}
\label{Sec_Results}

\subsection{Maintaining the Classification Accuracy}
\label{Sec_Results_Accuracy}

\begin{figure*}[t]
\centering
\includegraphics[width=\linewidth]{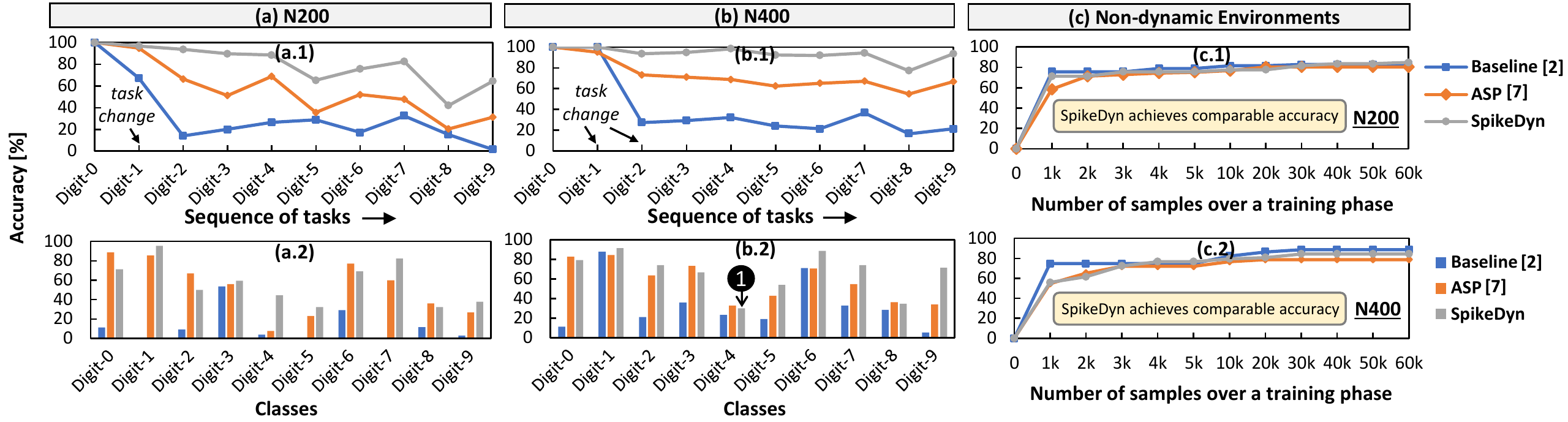}
\vspace{-0.7cm}
\caption{Accuracy in the \textit{dynamic environments}:
for most recently learned task in
(a.1) N200, (b.1) N400; and for the previously learned tasks in (a.2) N200, (b.2) N400. 
Accuracy in the \textit{non-dynamic environments}: over the presentation of training samples in (c.1) N200 and (c.2) N400.}   
\label{Fig_Results_Accuracy}
\vspace{-0.5cm}
\end{figure*}

\begin{figure}[hbtp]
\centering
\includegraphics[width=\linewidth]{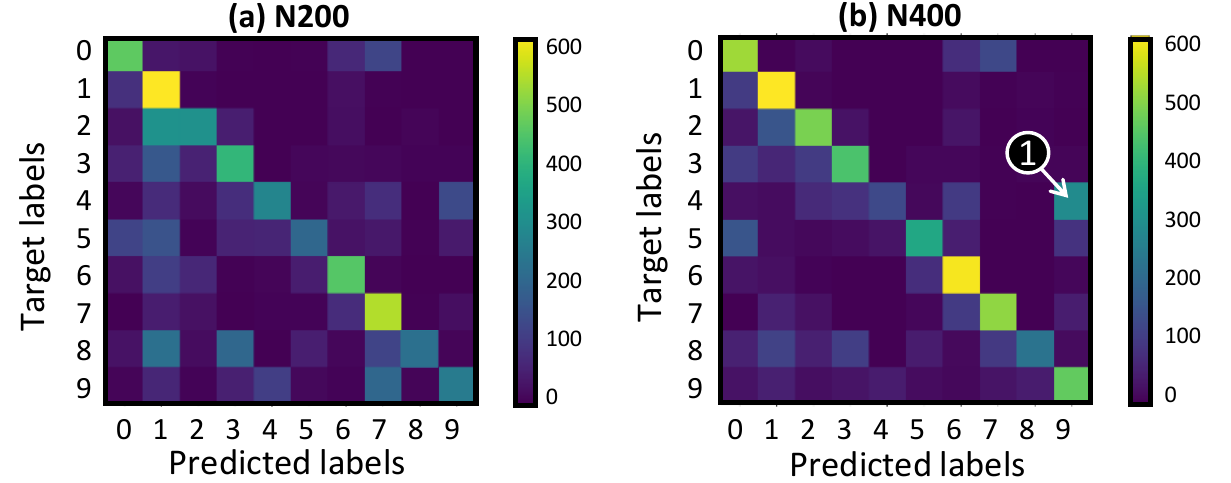}
\vspace{-0.6cm}
\caption{Confusion matrices of the SpikeDyn for classifying the previously learned tasks, which show the relation between the target labels and the predicted labels in (a) N200 and (b) N400.}   
\label{Fig_Results_ConfuseMatrix}
\vspace{-0.3cm}
\end{figure}

\textbf{Dynamic Environments:}
We evaluated the classification accuracy for two cases, i.e., (1) classifying the most recently learned task, which represent the capability of learning new task; and (2) classifying the previously learned tasks, which represent the capability of retaining old information. 

\textit{Case}-1: Figs.~\ref{Fig_Results_Accuracy}(a.1) and \ref{Fig_Results_Accuracy}(b.1) show the accuracy when the network classifies the most recently learned task (or learning new task), for N200 and N400 respectively.
The ASP achieves better accuracy than the baseline since the ASP employs adaptive learning rate and weight decay, while the baseline does not consider the dynamic tasks in its learning.
Here, our SpikeDyn improves the learning capabilities more than the ASP. 
The SpikeDyn improves the accuracy by up to 38\% (avg. 23\%) than the ASP for N200, and by up to 29\% (avg. 21\%) for N400.
The reason is that, the SpikeDyn employs: 
(1) a more careful mechanism to determine the rates of weight potentiation and depression for learning new features, 
(2) an appropriate threshold potential to adjust some neurons to be active in the learning process, 
(3) weight decay rate that effectively removes the old and insignificant information, and 
(4) reduction of the spurious weight updates.

\textit{Case}-2: 
Figs.~\ref{Fig_Results_Accuracy}(a.2) and \ref{Fig_Results_Accuracy}(b.2) show the accuracy when the network classifies the previously learned tasks (or retaining old information), for N200 and N400 respectively.
The baseline performs the worst as it does not decrease the weights, thereby suffering from mixed information in its synapses. 
Here, our SpikeDyn shows comparable accuracy to the ASP.
The SpikeDyn improves the accuracy by up to 36\% (avg. 4\%) than the ASP for N200, and by up to 37\% (avg. 8\%) for N400.
The reason is that, the SpikeDyn employs: 
(1) a threshold potential that tunes some neurons to be inactive in the learning process, thereby retaining the old yet significant information, and 
(2) weight decay rate that does not remove the old yet significant connections.
Furthermore, we also observe that, some classes are relatively difficult to learn in dynamic scenarios, especially in the case of retaining old information.  
For instance, in N400 case, the accuracy for classifying digit-4 is low, as indicated by label-\circledB{1} in Fig.~\ref{Fig_Results_Accuracy}(b.2). 
It is because a considerable number of misclassification happens when the digit-4 is recognized as another digit (i.e., digit-9), as indicated by label-\circledB{1} in Fig.~\ref{Fig_Results_ConfuseMatrix}(b). 
The reason is that, the learned features from digit-4 are gradually changed to represent the features of digit-9 over a training period, due to their overlapped features and the sequence of learning tasks. 
Therefore, some neurons that recognize digit-4 at the early of the training period, are changed to recognize digit-9 at the end of the training period. 

\textbf{Non-dynamic Environments:}
Figs.~\ref{Fig_Results_Accuracy}(c.1) and \ref{Fig_Results_Accuracy}(c.2) show the accuracy over the presentation of training samples for N200 and N400, respectively.  
The results show that our SpikeDyn achieves comparable accuracy to other techniques. 
The reason is that, our SpikeDyn employs effective learning rates to potentiate and decrease the weights, while reducing the spurious updates. 
In this manner, each weight update is adjusted appropriately, and hence the accuracy is maintained.
Such observations are important, as SNN systems may have a different number of training samples available from the environment. 
Thus, the users can devise a strategy to define the minimum training samples for achieving the targeted accuracy.


\begin{figure}[t]
\centering
\includegraphics[width=\linewidth]{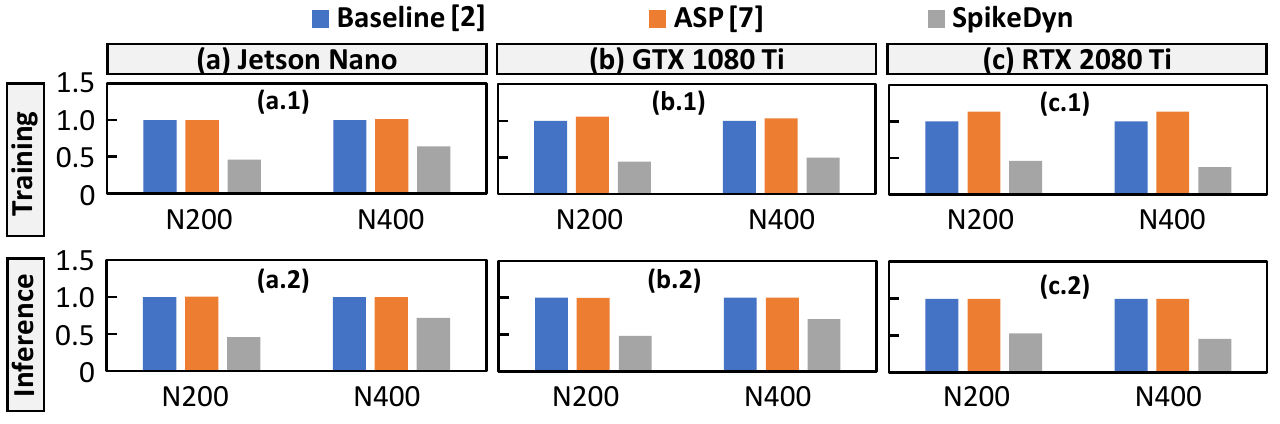}
\vspace{-0.6cm}
\caption{The energy consumption (normalized to the baseline) for training and inference phases, and across different sizes of networks and different GPUs.} 
\label{Fig_Results_Energy}
\end{figure}

\subsection{Reduction of the Energy Consumption}
\label{Sec_Results_Energy}

Fig.~\ref{Fig_Results_Energy} shows that SpikeDyn reduces the energy consumption compared to other techniques, across different sizes of network and different GPUs, for both the dynamic and non-dynamic environments. 
For N200, our SpikeDyn reduces the energy consumption from the ASP by up to 59\% (avg. 57\%) for training, and by up to 54\% (avg. 51\%) for inference. 
For N400, our SpikeDyn reduces the energy consumption from the ASP by up to 66\% (avg. 51\%) for training, and by up to 54\% (avg. 37\%) for inference.
The energy savings in training come from the elimination of the inhibitory neurons, the reduction of spurious updates and exponential calculations.
Meanwhile, the energy savings in inference mainly come from the elimination of inhibitory neurons.
Furthermore, we also observe the processing time of the SpikeDyn for training and inference phases (see Table~\ref{Table_ProcTimeGPUs}).
The results show that running an SNN model on the Embedded GPU (Jetson Nano) requires a longer time than the GPGPUs, as the Embedded GPU has less number of cores, memory size and bandwidth. 
Therefore, the users should devise a strategy for defining the number of samples in the training and inference phases, to comply with the use cases' requirements, especially on the embedded applications.

\section{Conclusion}
\label{Sec_Conclusion}

We propose a novel SpikeDyn framework that supports continual and unsupervised learning for SNNs, while reducing the energy consumption, by optimizing of SNN operations and improving the learning algorithm. 
The experimental results show that our SpikeDyn incurs less energy and improves the accuracy, as compared to the state-of-the-art in both the dynamic and non-dynamic scenarios. 
Therefore, our SpikeDyn would enable an energy-efficient embedded SNNs with one-time deployment.

\begin{table}[t]
	\caption{Processing Time of SpikeDyn on \textit{Full} MNIST Dataset (in hours).}
	\label{Table_ProcTimeGPUs}
	\vspace{-0.2cm}
	\centering
	\scriptsize
	\begin{tabular}{|c|c|c|c|c|c|c|}
		\hline 
		\multirow{2}{*}{\textbf{Process}} & \multicolumn{2}{c|}{\textbf{Jetson Nano}} & \multicolumn{2}{c|}{\textbf{GTX 1080 Ti}} & \multicolumn{2}{c|}{\textbf{RTX 2080 Ti}} \\
		\cline{2-7}
		& \textbf{N200} & \textbf{N400} & \textbf{N200} & \textbf{N400} & \textbf{N200} & \textbf{N400} \\
		\hline
		\hline
		Training & 35.0 & 36.3 & 5.0 & 5.3 & 3.9 & 4.1 \\
		\hline
		Inference & 4.7 & 4.8 & 0.7 & 0.7 & 0.6 & 0.6 \\
		\hline
		Inference of an & \multirow{2}{*}{1.71s} & \multirow{2}{*}{1.74s} & \multirow{2}{*}{0.25s} & \multirow{2}{*}{0.25s} & \multirow{2}{*}{0.2s} & \multirow{2}{*}{0.2s} \\
		image (seconds) & & & & & & \\
		\hline
	\end{tabular}
\end{table}

\vspace{-0.1cm}
\section{Acknowledgment}
\vspace{-0.1cm}
This work was partly supported by the Indonesia Endowment Fund for Education (IEFE/LPDP) Scholarship Program from Ministry of Finance, Indonesia.

\bibliographystyle{IEEEtran}
\bibliography{bibliography}

\begin{thebibliography}{10}
\providecommand{\url}[1]{#1}
\csname url@samestyle\endcsname
\providecommand{\newblock}{\relax}
\providecommand{\bibinfo}[2]{#2}
\providecommand{\BIBentrySTDinterwordspacing}{\spaceskip=0pt\relax}
\providecommand{\BIBentryALTinterwordstretchfactor}{4}
\providecommand{\BIBentryALTinterwordspacing}{\spaceskip=\fontdimen2\font plus
\BIBentryALTinterwordstretchfactor\fontdimen3\font minus
  \fontdimen4\font\relax}
\providecommand{\BIBforeignlanguage}[2]{{%
\expandafter\ifx\csname l@#1\endcsname\relax
\typeout{** WARNING: IEEEtran.bst: No hyphenation pattern has been}%
\typeout{** loaded for the language `#1'. Using the pattern for}%
\typeout{** the default language instead.}%
\else
\language=\csname l@#1\endcsname
\fi
#2}}
\providecommand{\BIBdecl}{\relax}
\BIBdecl

\bibitem{Ref_Pfeiffer_DLSNN_FNINS18}
M.~Pfeiffer and T.~Pfeil, ``Deep learning with spiking neurons: Opportunities
  and challenges,'' \emph{Frontiers in Neuroscience}, vol.~12, 2018.

\bibitem{Ref_Diehl_STDPmnist_FNCOM15}
P.~Diehl and M.~Cook, ``Unsupervised learning of digit recognition using
  spike-timing-dependent plasticity,'' \emph{Frontiers in Computational
  Neuroscience}, vol.~9, p.~99, 2015.

\bibitem{Ref_Srinivasan_EnhPlast_IJCNN17}
G.~{Srinivasan} \emph{et~al.}, ``Spike timing dependent plasticity based
  enhanced self-learning for efficient pattern recognition in spiking neural
  networks,'' in \emph{Proc. of IJCNN}, May 2017, pp. 1847--1854.

\bibitem{Ref_Hazan_SOMSNN_IJCNN18}
H.~{Hazan} \emph{et~al.}, ``Unsupervised learning with self-organizing spiking
  neural networks,'' in \emph{Proc. of IJCNN}, July 2018, pp. 1--6.

\bibitem{Ref_Saunders_LCSNN_NeuNet19}
D.~J. Saunders \emph{et~al.}, ``Locally connected spiking neural networks for
  unsupervised feature learning,'' \emph{Neural Networks}, vol. 119, 2019.

\bibitem{Ref_Putra_FSpiNN_TCAD20}
R.~V.~W. {Putra} and M.~{Shafique}, ``Fspinn: An optimization framework for
  memory-efficient and energy-efficient spiking neural networks,'' \emph{IEEE
  TCAD}, vol.~39, no.~11, pp. 3601--3613, 2020.

\bibitem{Ref_Panda_ASP_JETCAS18}
P.~{Panda} \emph{et~al.}, ``Asp: Learning to forget with adaptive synaptic
  plasticity in spiking neural networks,'' \emph{IEEE JETCAS}, vol.~8, no.~1,
  March 2018.

\bibitem{Ref_Lobo_SNNonline_NeuNet19}
J.~L. Lobo \emph{et~al.}, ``Spiking neural networks and online learning: An
  overview and perspectives,'' \emph{Neural Networks}, vol. 121, 2020.

\bibitem{Ref_Lesort_CLrobot_IF20}
T.~Lesort \emph{et~al.}, ``Continual learning for robotics: Definition,
  framework, learning strategies, opportunities and challenges,''
  \emph{Information Fusion}, vol.~58, pp. 52--68, 2020.

\bibitem{Ref_Anthes_LifelongLearn_ACM19}
G.~Anthes, ``Lifelong learning in artificial neural networks,'' \emph{Commun.
  ACM}, vol.~62, no.~6, p. 13–15, May 2019.

\bibitem{Ref_Ven_BIReplay_Nature20}
G.~M. van~de Ven \emph{et~al.}, ``Brain-inspired replay for continual learning
  with artificial neural networks,'' \emph{Nature communications}, vol.~11,
  no.~1, pp. 1--14, 2020.

\bibitem{Ref_Allred_ForcedFiring_IJCNN16}
J.~M. {Allred} and K.~{Roy}, ``Unsupervised incremental stdp learning using
  forced firing of dormant or idle neurons,'' in \emph{Proc. of IJCNN}, 2016,
  pp. 2492--2499.

\bibitem{Ref_Chen_Lifelong_MLP18}
Z.~Chen and B.~Liu, ``Lifelong machine learning,'' \emph{Synthesis Lectures on
  Artificial Intelligence and Machine Learning}, vol.~12, no.~3, 2018.

\bibitem{Ref_Parisi_CLL_NeuNet19}
G.~I. Parisi \emph{et~al.}, ``Continual lifelong learning with neural networks:
  A review,'' \emph{Neural Networks}, vol. 113, pp. 54 -- 71, 2019.

\bibitem{Ref_McCloskey_CI_Elsevier89}
M.~McCloskey and N.~J. Cohen, ``Catastrophic interference in connectionist
  networks: The sequential learning problem,'' \emph{Psychology of Learning and
  Motivation}, vol.~24, pp. 109 -- 165, 1989.

\bibitem{Ref_Kirkpatrick_EWC_PNAS17}
J.~Kirkpatrick \emph{et~al.}, ``Overcoming catastrophic forgetting in neural
  networks,'' \emph{PNAS}, vol. 114, no.~13, pp. 3521--3526, 2017.

\bibitem{Ref_Lee_OvercomeCF_NIPS17}
S.-W. Lee \emph{et~al.}, ``Overcoming catastrophic forgetting by incremental
  moment matching,'' in \emph{Proc. of NIPS}, 2017, pp. 4652--4662.

\bibitem{Ref_Wysoski_OLstructural_ICANN06}
S.~G. Wysoski \emph{et~al.}, ``On-line learning with structural adaptation in a
  network of spiking neurons for visual pattern recognition,'' in \emph{Proc.
  of ICANN}, 2006, pp. 61--70.

\bibitem{Ref_Allred_CFN_FNINS20}
J.~M. Allred and K.~Roy, ``Controlled forgetting: Targeted stimulation and
  dopaminergic plasticity modulation for unsupervised lifelong learning in
  spiking neural networks,'' \emph{Frontiers in Neuroscience}, vol.~14, p.~7,
  2020.

\bibitem{Ref_Tavanaei_DLSNN_Neunet18}
A.~Tavanaei \emph{et~al.}, ``Deep learning in spiking neural networks,''
  \emph{Neural Networks}, vol. 111, pp. 47--63, 2019.

\bibitem{Ref_Gautrais_SpikeCoding_Bio98}
J.~Gautrais and S.~Thorpe, ``Rate coding versus temporal order coding: a
  theoretical approach,'' \emph{Biosystems}, vol.~48, no.~1, pp. 57--65, 1998.

\bibitem{Ref_Kayser_PhaseCoding_Neuron09}
C.~Kayser \emph{et~al.}, ``Spike-phase coding boosts and stabilizes information
  carried by spatial and temporal spike patterns,'' \emph{Neuron}, vol.~61,
  no.~4, pp. 597 -- 608, 2009.

\bibitem{Ref_Thorpe_RankOrder_Springer98}
S.~Thorpe and J.~Gautrais, ``Rank order coding,'' in \emph{Computational
  neuroscience}.\hskip 1em plus 0.5em minus 0.4em\relax Springer, 1998, pp.
  113--118.

\bibitem{Ref_Park_BurstSNN_DAC19}
S.~Park \emph{et~al.}, ``Fast and efficient information transmission with burst
  spikes in deep spiking neural networks,'' in \emph{Proc. of DAC}, 2019,
  p.~53.

\bibitem{Ref_Hazan_BindsNET_FNINF18}
H.~Hazan \emph{et~al.}, ``Bindsnet: A machine learning-oriented spiking neural
  networks library in python,'' \emph{Frontiers in Neuroinformatics}, vol.~12,
  2018.

\bibitem{Ref_Han_DeepCompress_ICLR16}
S.~Han \emph{et~al.}, ``Deep compression: Compressing deep neural networks with
  pruning, trained quantization and huffman coding,'' \emph{arXiv preprint
  arXiv:1510.00149}, 2015.

\end{thebibliography}
\end{document}